\documentclass[letterpaper, 10 pt, conference]{ieeeconf}  % Comment this line out if you need a4paper

\usepackage{algorithm}
\usepackage{algorithmic}
\usepackage{amsmath}
\usepackage{amssymb}
\usepackage{accents}
\usepackage{bm}
\usepackage{graphicx}
\usepackage{booktabs}
\usepackage{multirow}
\usepackage{rotating}
\usepackage{hyperref}
\usepackage{cleveref}
\usepackage{subcaption}
\usepackage{caption}
\usepackage{array}
\usepackage{dblfloatfix}

\Crefname{figure}{Fig.}{Figs.}
\crefname{figure}{Fig.}{Figs.}
\Crefname{table}{Tab.}{Tabs.}
\crefname{table}{Tab.}{Tabs.}
\Crefname{equation}{Eq.}{Eqs.}
\crefname{equation}{Eq.}{Eqs.}

\usepackage{amsmath,amsfonts,bm}

\def\eqref#1{equation~\ref{#1}}
\def\1{\bm{1}}

\def\vtheta{{\bm{\theta}}}
\def\va{{\bm{a}}}

\def\vc{{\bm{c}}}

\def\vh{{\bm{h}}}

\def\vp{{\bm{p}}}

\def\vv{{\bm{v}}}

\def\vx{{\bm{x}}}
\def\vy{{\bm{y}}}
\def\vz{{\bm{z}}}

\DeclareMathAlphabet{\mathsfit}{\encodingdefault}{\sfdefault}{m}{sl}
\SetMathAlphabet{\mathsfit}{bold}{\encodingdefault}{\sfdefault}{bx}{n}

\def\gD{{\mathcal{D}}}

\def\gX{{\mathcal{X}}}
\def\gY{{\mathcal{Y}}}

\newcommand{\E}{\mathbb{E}}
\newcommand{\Ls}{\mathcal{L}}

\IEEEoverridecommandlockouts                              % This command is only needed if 
\title{\LARGE \bf Active Neural Mapping at Scale}

\author{Zijia Kuang$^{1}$, Zike Yan$^{1,\dag}$, Hao Zhao$^{1}$, Guyue Zhou$^{1}$, and Hongbin Zha$^{2}$% <-this % stops a space
\thanks{\dag~Corresponding author. {\tt\small yanzike@air.tsinghua.edu.cn}}% <-this % stops a space
\thanks{$^{1}$ Zijia Kuang, Zike Yan, Hao Zhao, and Guyue Zhou are with the Institute for AI Industry Research (AIR), Tsinghua University, Beijing, China.
        }%
\thanks{$^{2}$ Hongbin Zha is with the School of Intelligence Science and Technology, Peking University, Beijing, China}
}

\begin{document}

\maketitle
\thispagestyle{empty}
\pagestyle{empty}

%%%%%%%%%%%%%%%%%%%%%%%%%%%%%%%%%%%%%%%%%%%%%%%%%%%%%%%%%%%%%%%%%%%%%%%%%%%%%%%%
\begin{abstract}
We introduce a NeRF-based active mapping system that enables efficient and robust exploration of large-scale indoor environments. The key to our approach is the extraction of a generalized Voronoi graph (GVG) from the continually updated neural map, leading to the synergistic integration of scene geometry, appearance, topology, and uncertainty. Anchoring uncertain areas induced by the neural map to the vertices of GVG allows the exploration to undergo adaptive granularity along a safe path that traverses unknown areas efficiently. Harnessing a modern hybrid NeRF representation, the proposed system achieves competitive results in terms of reconstruction accuracy, coverage completeness, and exploration efficiency even when scaling up to large indoor environments. Extensive results at different scales validate the efficacy of the proposed system.
\end{abstract}

\section{Introduction}
\label{sec:intro}
Accurate modeling of the surrounding environment for an embodied agent is of great importance towards spatial intelligence. Recent pivotal advances in the relevant fields are the developments of implicit neural representations (INRs)~\cite{Sitzmann2020nips,Mildenhall2020eccv, Muller2022tog, Yuce2022cvpr}. Scene reconstruction is formulated as a coordinate-based low-dimensional regression problem and solved through gradient-based optimization, where a compact model can be attained to recover accurate scene geometry~\cite{Lindell2022cvpr}, appearance~\cite{Mildenhall2020eccv,Barron2023iccv}, and semantics~\cite{Zhi2021iccv,Kerr2023iccv}. Nevertheless, the inherent under-determined nature as an inverse problem leads to susceptibility to artifacts, as the continuous scene representation is inferred from incomplete observations in an unknown environment.

\begin{figure}[t]
 	\centering
 	\includegraphics[width=0.99\linewidth]{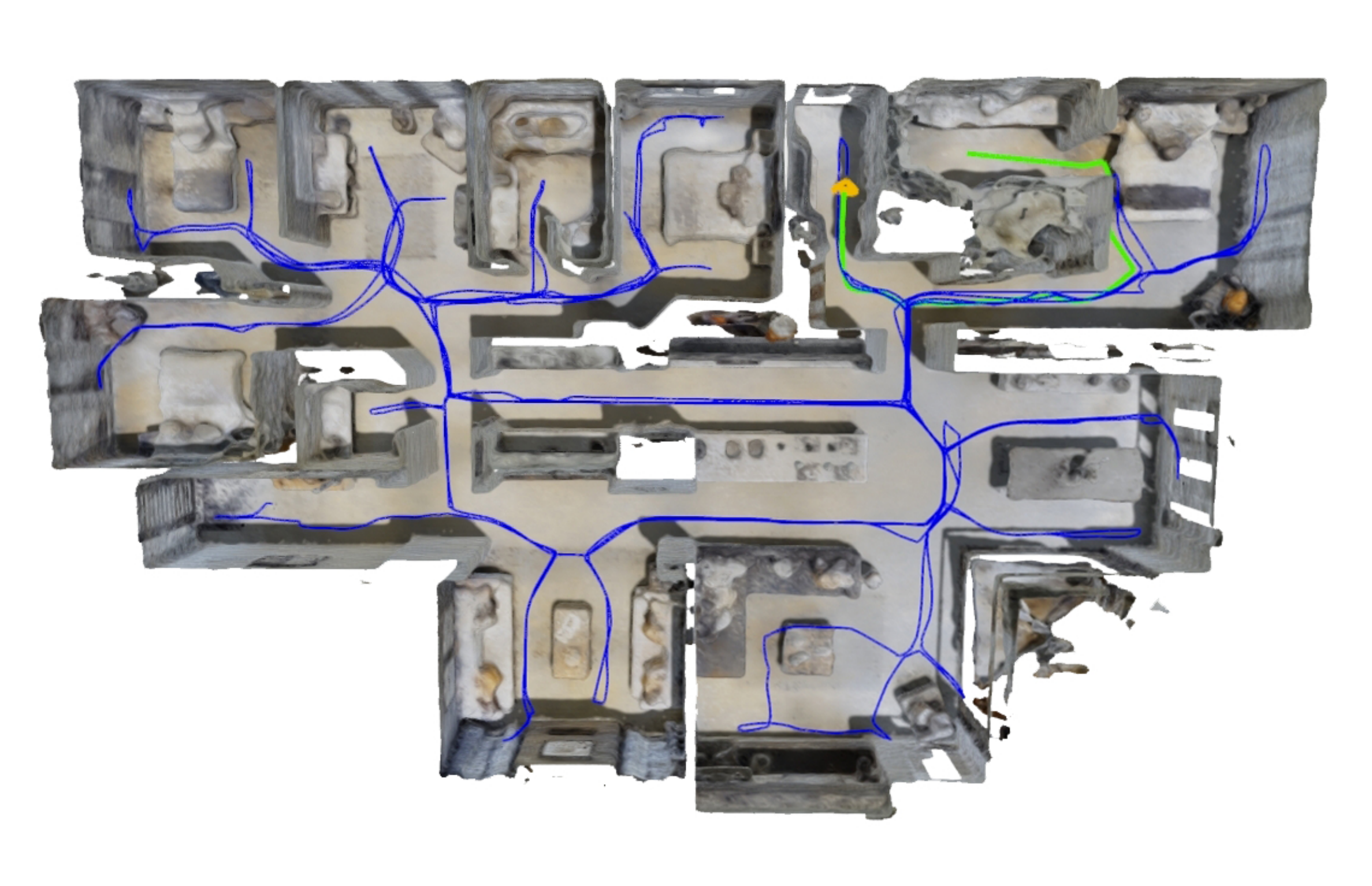}	
\caption{The reconstructed mesh through autonomous exploration with a continually-learned neural map (MP3D-zsNo4~\cite{Matterport2017_3DV} with 23 rooms). The synergistic integration of geometry, appearance, topology, and uncertainty information within the neural map leads to complete and accurate reconstruction of large-scale environments.}
    \label{fig:teaser}
\end{figure}

Note that the ill-posed issue not only applies to the differentiable optimization of INRs, but also to the conventional data fusion paradigms based on discretization-based representations such as volume grids and meshes. Insufficient observations lead to incompleteness and thus require autonomous exploration and reconstruction of the environment, or namely active mapping, an area that has been studied for decades~\cite{Connolly1985icra}. While the problem is well-formulated by approaching the surface frontiers~\cite{Yamauchi1997frontier} or by selecting the next-best-view samples through certain criteria~\cite{Bircher2016icra}, we aim to address in this work the following question: \emph{Is information within a neural map sufficient for fast and thorough exploration in an unknown indoor environment?}

As defined in~\cite{Yamauchi1997frontier}, exploration is "the act of moving through an unknown environment while building a map for subsequent navigation". That is to say, the neural map should be updated on-the-fly, quantify uncertainty beyond past observations, and provide a collision-free path to traverse the environment. This is formulated in~\cite{Yan2023iccv}, namely "active neural mapping", in a greedy fashion. Implicit neural representations have been challenged regarding the slow convergence, the lack of interpretability, and the absence of explicit structural information. While recent advances in implicit neural representation allow efficient online reconstruction~\cite{Wang2023cvpr,Sandstrom2023iccv} and reasonable uncertainty quantification~\cite{Yan2023iccv, Goli2023bayes}, we argue that the lack of explicit structural information prevents INRs from interpreting granularity at different levels. This is one typical gap between neural and symbolic representations and hinders efficient exploration with NeRF-wise representations.

In this paper, we propose to extract a generalized Voronoi graph (GVG) from the neural map to organize the information at different levels of detail in a structured manner. Edges of GVG delineate the spatial partitions of surface points according to Voronoi tessellation, where the inherent sparsity and the focus on maximizing clearance allow efficient and safe path planning for autonomous exploration (see~\cref{fig:teaser}). As vertices of GVG symbolize the common boundaries of multiple sub-areas, they encapsulate high geometry complexity and serve as critical decision points. Therefore, we anchor the fine-grained details from the neural map to Voronoi vertices to balance between efficient exploration and information utilization. The exploration can then be conducted under adaptive granularity through information-guided traversal of Voronoi vertices recursively. 

$\bullet$ We introduce a NeRF-based active mapping system that can explore large-scale indoor environments with up to 20+ rooms comprehensively.

$\bullet$ We extract structured information within the neural map using a generalized Voronoi graph to take both accessible areas of interest and visible areas of interest into account.

$\bullet$ We maintain a hierarchical framework to balance between information utilization and exploration efficiency with competitive performance.

\section{Related Work}
\label{sec:related_work}
%{\color{red}{Active vision}}
\paragraph{Autonomous exploration and reconstruction}
Autonomous exploration and reconstruction reduce the uncertainty of the map by finding the areas to be explored iteratively. Frontier-based methods maintain the boundary between explored and unvisited space~\cite{Yamauchi1997frontier} and gradually expand the coverage. This is usually achieved through mapping, frontier detection, and selection, where different strategies are made to extend the method to multi-robot scenario~\cite{Dong2019tog,Kai2022cvpr} and 3D space~\cite{Shen2012icra,Zhou2021ral} with better completeness and accuracy. Another line of research tends to find the next-best-view in a receding horizon to maximize the target objective~\cite{Bircher2016icra}. As a sampling-based method, attempts are made to determine the per-sample information gain~\cite{Schmid2020ral,Dang2018icra} and connect these samples through feasible paths. 

There are also hybrid works~\cite{Selin2019ral,Batinovic2022ral} that trade-off between local accuracy and global completeness for efficient and accurate exploration. Recent methods focus on reward-driven learning policy instead of heuristic designs. This is usually achieved in a modular~\cite{Chaplot2019iclr,Gervet2023sr} or end-to-end~\cite{Chen2019iclr,Sridhar2023nomad} fashion. The learned prior can also predict occupancy in unseen areas ~\cite{Ramakrishnan2020eccv} to accelerate the exploration. We, on the other hand, exploit the expressive representation power of neural maps to guide agent movement using fail-safe graph-based planning strategy.

\paragraph{NeRF-based SLAM}
The differentiable rendering fashion through a compact but expressive neural network~\cite{Mildenhall2020eccv} has been commonly adopted as one promising method for 3D reconstruction. The method is later extended to continual learning based solutions~\cite{Yan2021iccv, Sucar2021iccv} to handle constant distribution shifts from sequential observations. Keyframes are restored as replayed buffers to constantly constrain the optimization. The computational efficiency and global accuracy are further enhanced through different neural representations~\cite{Zhu2022cvpr, Sandstrom2023iccv, Wang2023cvpr} and global optimization~\cite{Zhang2023iccv}.

One critical issue for NeRF-based scene reconstruction is the ill-posed nature as an inverse problem~\cite{Sitzmann2020nips}. Insufficient observations will lead to artifacts and prediction errors. Different methods tackle the problem in an offline~\cite{Pan2022eccv} or online~\cite{Ran2023ral} setting. Recent studies~\cite{Yan2023iccv, Feng2024cvpr} extend the INR-based active mapping problem to an online scene-scale level in a greedy fashion. In this paper, we further exploit the structure information within the neural map to accelerate exploration with promising reconstruction accuracy.
\section{Active Neural Mapping with NeRF}
\label{sec:pre}

\begin{figure*}
        \vspace{1em}
        \centering
	\subcaptionbox{Top-down map\label{subfig:topdown}}[0.32\textwidth]
	{\includegraphics[width=0.32\textwidth]{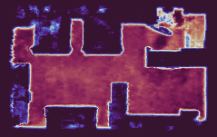}}
        \hfill
        \centering
	\subcaptionbox{Generalized Voronoi graph\label{subfig:voronoi}}[0.32\textwidth]
	{\includegraphics[width=0.32\textwidth]{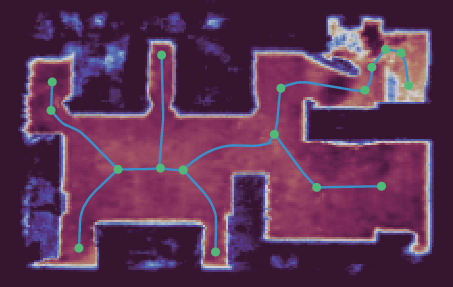}}
        \hfill
        \centering
	\subcaptionbox{Generalized Voronoi graph with ROIs\label{subfig:anchor}}[0.32\textwidth]
	{\includegraphics[width=0.32\textwidth]{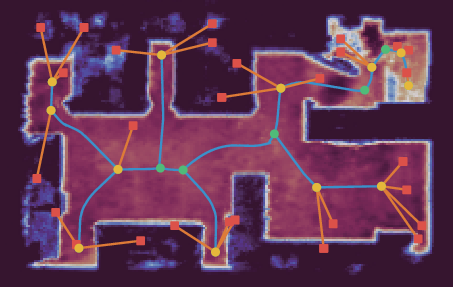}}
	\caption{The structuring process through Voronoi tessellation. The Generalized Voronoi graph (GVG) (b) is extracted as equidistant samples to the zero-crossing surfaces (a). The visible regions of interest (ROIs) guided by perturbed map parameters are anchored to neighboring vertices (c).}
	\label{fig:structuring} 
\end{figure*}
We target an active neural mapping~\cite{Yan2023iccv} problem, where neural map parameters $\vtheta^t$ and the action control $\va^t$ are optimized recursively to best explore and reconstruct an environment unknown \emph{a priori}. The problem can be solved by finding a sequence of accessible poses~
$\{\vv^t\}_{1:n} \in \mathrm{{\mathbf{SE}}}(2)$, such that the neural map $f(\vx;\vtheta): \gX \to \gY$ can be continually updated given new observations $\vz(\vv^t) \subset \gD$ to gradually minimize the reconstruction error $\E_{(\vx,\vy)\sim\gD}(\Ls(\vx,\vy; \vtheta))$. 
To maintain the high-fidelity geometry and appearance of the scene, we optimize an implicit representation $f(\vx;\vtheta)\rightarrow(\mathbf{c}, s)$ that maps 3D spatial coordinates to color $\mathbf{c}$ and truncated signed distance (TSDF) values $s$~\cite{Wang2023cvpr}. The compact representation allows efficient query of coordinate values through a single forward pass. Depth and color images can be synthesized through volume rendering~\cite{Sucar2021iccv, Zhu2022cvpr, Wang2023cvpr} as the weighted sum of all samples $\mathbf{p}_i = \mathbf{o}+d_i\mathbf{r}, i\in \{1,\cdots,N\}$ along a ray $\mathbf{r}(u,v)$:
\begin{equation}
	\label{eq:volume_rendering}
	\hat{\mathbf{c}}[u,v] = \frac{1}{\sum_{i=1}^{N} w_i} \sum_{i=1}^{N} w_i \mathbf{c}_i, 
        \hat{d}[u,v] = \frac{1}{\sum_{i=1}^{N} w_i} \sum_{i=1}^{N} w_i d_i,
\end{equation}
where the weight $\omega_i$ is the multiplication of two Sigmoid functions with a truncation threshold $\lambda_{tr}$ as:
\begin{equation}
	\label{eq:mapping}
	\omega_i = \sigma\left(\frac{s_i}{\lambda_{tr}}\right) \sigma\left(-\frac{s_i}{\lambda_{tr}}\right).
\end{equation}

Though the neural map provides continuous and fine-grained details of the environment, enumerating the entire cascade of decisions $\{\mathbf{a}\}_{1:n}$ in the working space without \emph{a priori} is non-trivial. Central to this work is a structured representation $\{\mathcal{V}, \mathcal{E}\}$ that manages information within the neural map $f(\vx;\vtheta)$ and enforces efficient and thorough exploration of unknown environments.

\subsection{Structuring NeRF into a navigational map}
To capture the essential navigational information within the network, we adopt Voronoi tessellation to divide the working space into sub-regions and capture the topology of the partially-observed accessible regions. The Voronoi graph itself is a roadmap of the free space~\cite{Choset2001topological}, where edges $\mathcal{E}$ define a set of accessible and connected waypoints that are equidistant to the scene surface, and vertices $\mathcal{V}$ are the meet points that the edges terminate at. As all surface points can be assigned to the sparse GVG edges according to Voronoi tessellation, the graph structures the information within a neural map into a compact and sparse form.

Denote a distance function from coordinate $\vx$ to a surface point $\vp_i$ as $d_i(\vx) = dist(\vp_i,\vx)$. The Voronoi tessellation defines the sub-region $S_i$ affected by a surface point $\vp_i$ as the set of points in the space that are closer to $\vp_i$ than to any other surface points as:
\begin{equation}
	\label{eq:voronoi_diagram}
	S_i = \{\vx| d_i(\vx) < d_j(\vx) \text{ for all } j \neq i\}.
\end{equation}

Note that the surface points $\{\vp_i\}_{1:m}$ can be efficiently queried from the neural map as zero-crossings\footnote{A top-down map is created in the horizontal plane as illustrated in Fig.~\ref{subfig:topdown}. For each grid point in the top-down map, we take the minimum value of the queried SDF given samples along the vertical direction.}. A Voronoi graph can be extracted~\cite{Choset2001topological} by first finding points equidistant to at least two surface points as edges, and then forming vertices as the intersections of multiple Voronoi edges. This can be done directly using the Voronoi tessellation by examining the edges of the polygons in the tessellation. We remove the vertices with negative distance values through forward passes. As illustrated in~\cref{subfig:voronoi}, the Voronoi graph derives explicit topology of the partially observed free space that greatly simplifies the planning problem with a sparse and discrete structure.

\subsection{Identifying the region of interests}
Given the topological structure of the accessible area, a path can be efficiently generated through graph-search algorithms once a target location is decided. With the partially observed topology and scene radiance from the instant Voronoi graph $\{\mathcal{V}, \mathcal{E}\}^t$ and neural map $f(\vx;\vtheta^t)$, we can take both accessible and visible region of interests into account for thorough exploration and accurate reconstruction.

To evaluate the reconstruction quality at a fine-grained level, we need to conduct uncertainty quantification from the continually learned neural map. Recent study~\cite{Yan2023iccv} indicates that loss landscapes are divergent when evaluating prediction error at different areas: the areas in the working space that lack constraints would lead to unstable minima, where small perturbations on the network parameters would result in evident prediction variation; the areas with constant supervision during the continual learning process would lead to a flat basin that is robust against perturbation. Note that the adopted network simultaneously outputs the geometry $s$ and the appearance $\vc$ of the scene. Perturbing different decoders leads to different selections of interest. The test-time reparameterization would not affect the performance of the adopted backbone and is amenable to new advances. We perturb the parameters of both the geometry and appearance decoder of the neural field with Gaussian noises and quantify the prediction variances given zero-crossing coordinates through multiple passes, where the surface points with higher variances will be considered as the visible area of interest in terms of both geometry and appearance:
\begin{equation}
	\vx = \arg\max \mathbb{V}_{\hat{\vtheta}\sim\mathit{N}(\vtheta^t,b^2\mathit{I})}[f(\mathbf{x};\hat{\vtheta})].
	\label{eq:variability}
\end{equation}

The visible region of interest inferred by the neural map is inherently dense and delicate. Geometric and photometric details would be frequently highlighted due to insufficient observations or training steps. Moreover, the target regions induced by the neural map may be non-traversable. Therefore, as illustrated in~\cref{subfig:anchor}, we cluster the visible regions of interest according to their geometric and topological information and anchor them to the near Voronoi vertices that serve as accessible regions of interest. We further depress the accessible region of interest if the Voronoi vertices fall into previously visited areas. Consequently, Voronoi vertices that are anchored with visible regions of interest or belong to non-visited areas would be activated as the target position candidates. This ensures a safe and traversable path recursively, where fine-grained observations can be achieved by adjusting the viewing angle towards the visible regions of interest once arriving at the selected vertex.

\subsection{Graph search for efficient planning}
As the informative cues within the neural map are anchored at the sparse Voronoi vertices and edges, efficient action decisions can be made given the agent pose and the selected Voronoi vertex by treating the problem as a single-source shortest path planning. Given the explicit topology of the free space, we employ the Dijkstra algorithm~\cite{dijkstra} to find the optimal path in the weighted graph, where the weights are the Euclidean distance between two adjacent vertices. The path can be found through an iterative process that involves selecting the neighboring vertex with a minimal cumulative distance from the current position. Replanning 
is conducted every time the agent approaches the target frustum. The sparse nature of the graph guarantees efficient planning in milliseconds.
\section{Implementations for Large-scale Scenes}
\label{sec:method}
Given the structured information, the complexity of areas to be visited greatly reduces to the sparse set of vertices within the Voronoi graph. Unlike the greedy strategy conducted in~\cite{Yan2023iccv} that is susceptible to local minima, the Voronoi graph assures completeness theoretically under certain conditions~\cite{Kim2010ar} and simplifies the search space to a graph structure without intersecting any obstacles. Despite the merits, the topology map based planning and the neural map based optimization need to be carefully designed to adapt to large-scale environments with satisfying efficiency and completeness. 

\subsection{Hierarchical framework with adaptive granularity}
Voronoi graph based planning, while advantageous for many reasons, also encounters specific drawbacks when deployed in large-scale complex scenes. As proved by~\cite{Aurenhammer1991voronoi,Kretzschmar2016ijrr}, given that the vertex on the graph has a minimum degree of three, the undirected graph satisfies:
\begin{equation}
	2|\mathcal{E}|=\sum_{v\in\mathcal{V}}deg(v)\geq 3|\mathcal{V}|.
	\label{eq:degree_of_vertices}
\end{equation}

Meanwhile, viewing Voronoi graph as a planar graph with the number of faces $l$ the same as the number of surface samples, Euler's formula leads to:
\begin{equation}
	|\mathcal{V}|-|\mathcal{E}|+l=2.
	\label{eq:euler_formula}
\end{equation}

Therefore, the graph grows linearly with the number of surface samples:
\begin{equation}
	|\mathcal{V}|\leq 3l, |\mathcal{E}|\leq 2l.
	\label{eq:linear_growing}
\end{equation}

\begin{figure}[t]
 	\centering
 	\includegraphics[width=0.99\linewidth]{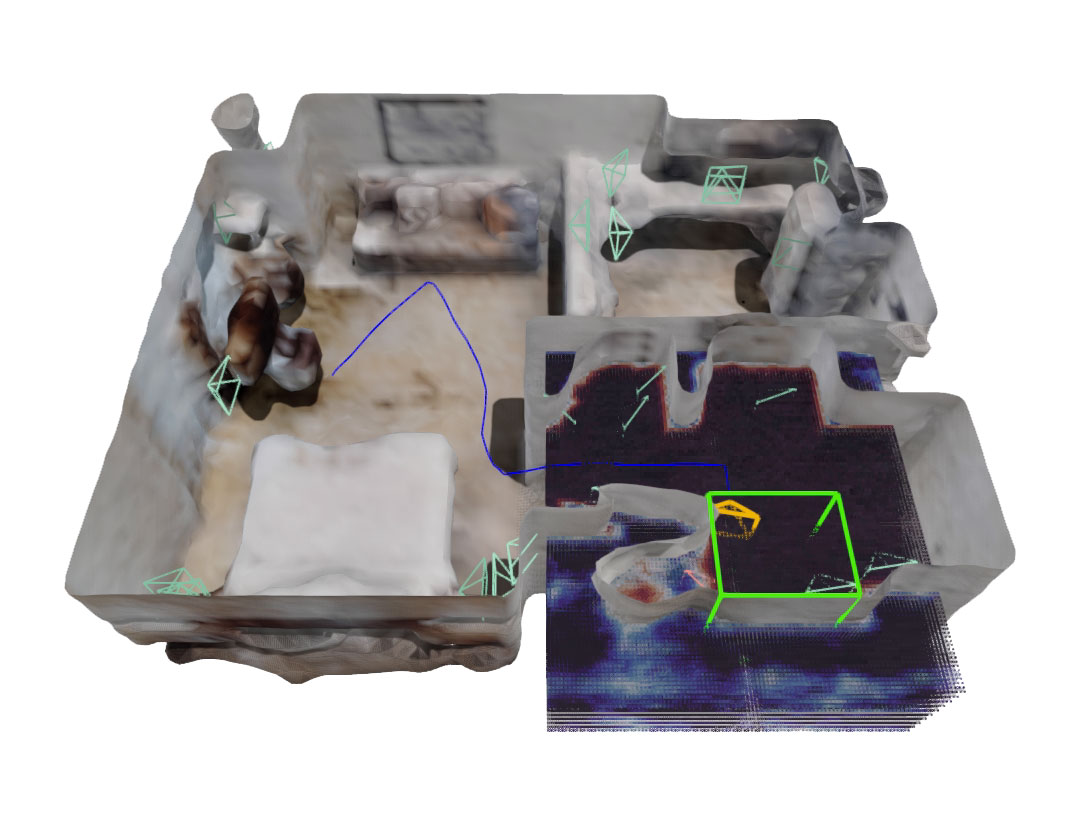}	
\caption{The fine-grained local horizon (the dark area near the frustum that indicates the nearby distance values) guarantees safe exploration, while the rest of the map maintains the global regions of interest coarsely.}
    \label{fig:local_horizon}
\end{figure}

Besides the extraction of the Voronoi graph, the graph-based search algorithm has a runtime complexity of $\mathcal{O}(|\mathcal{E}|+|\mathcal{V}|\log|\mathcal{V}|)$, leading to the complexity of $\mathcal{O}(l\log l)$. When encountering a scene with complex surface geometry, the rising number of zero-crossings will greatly increase the computational cost of the planning. To this end, we maintain a fixed resolution of the local horizon to extract precise zero-crossings, where coarse-level zero-crossings are extracted outside the local horizon. As illustrated in~\cref{fig:local_horizon}, the computational cost is upper-bounded while guaranteeing a safe path during exploration. This strategy greatly reduces the chances that the agent gets stuck by thin objects or near narrow zones while preserving promising efficiency. We prioritize anchored Voronoi vertices within the local horizon to avoid sudden maneuvers, and rotate toward the anchored regions of interest once arriving at the target position. Consequently, the agent tends to march quickly toward the selected Voronoi vertex and stay cautious at the critical decision points. The adaptive granularity for actively searching the informative observations achieves a nice balance between exploration efficiency and reconstruction accuracy.

\subsection{Hybrid neural representation}
The convergence of the neural representation greatly affects the exploration efficiency. As shown in~\cite{Yan2023iccv}, continual learning of a single multilayer perception (MLP) suffers from slow convergence, where the uncertainty of the neural map will guide the agent back to the visited areas with complex geometry details. In this paper, we adopt the joint coordinate and parametric encoding similar to Co-SLAM~\cite{Wang2023cvpr} to best tradeoff between fast convergence and accurate modeling. For scene geometry, the coordinate is encoded by the combination of a hash-encoded multi-resolution feature grid~\cite{Muller2022tog} $\{\mathcal{F}_{\alpha}^l(\vx)\}_{l=1}^{L}$ and a One-blob encoding $\gamma(\vx)$~\cite{Muller2019tog}, and decoded into the predicted TSDF value $s$ and a feature $\vh$. For scene radiance, the color is decoded given the concatenated features of $\vh$ and $\gamma(\vx)$:
\begin{equation}
	f_\tau(\gamma(\vx),\{\mathcal{F}_{\alpha}^l(\vx)\}_{l=1}^{L}) = \{\vh, s\},
	f_\phi(\gamma(\vx),\vh) = \vc.
	\label{eq:coslam_net}
\end{equation}

\begin{table*}[bt]
        \vspace{1em}
	\centering
	\caption{Comparison against relevant methods regarding the completeness (\%↑/$cm$↓) of actively-captured observations.}
	\label{tab:method_compare}
        \renewcommand\arraystretch{1.3}
        \setlength{\tabcolsep}{6mm}{
	\begin{tabular}{lccccc} 
		\toprule
		%\hline
		& \textbf{Random} & \textbf{FBE}~\cite{Yamauchi1997frontier} & \textbf{UPEN}~\cite{Georgakis2022icra} & \textbf{ANM}~\cite{Yan2023iccv} & \textbf{Ours} \\
		\toprule
		\textbf{Gibson}-Cantwell & 24.43/59.59 & 40.93/37.03 & 39.42/42.12 & 61.36/17.67 & \textbf{85.09}/\textbf{7.25}\\ 
		\hline
		\textbf{Gibson}-Denmark & 27.83/50.42 & 70.28/12.40 & 66.41/17.34 & 85.86/3.78 & \textbf{91.85}/\textbf{2.00}\\ 
		\hline
		\textbf{Gibson}-Eastville & 14.32/72.39 & 58.49/24.08 & 51.51/28.16 & 74.21/11.36 & \textbf{87.66}/\textbf{6.82}\\ 
		\hline
		\textbf{Gibson}-Elmira & 66.29/11.63 & 72.69/10.40 & 82.14/5.35 & 91.65/2.57 & \textbf{95.42}/\textbf{1.40}\\ 
		\hline
		\textbf{Gibson}-Eudora & 53.89/23.24 & 76.65/8.11 & 75.74/9.18 & 90.12/2.27 & \textbf{93.94}/\textbf{1.49}\\ 
		\hline
		\textbf{Gibson}-Greigsville & 75.44/6.97 & 90.34/2.62 & 73.72/16.34 & 92.47/1.78 & \textbf{98.65}/\textbf{0.64}\\ 
		\hline
		\textbf{Gibson}-Pablo & 46.87/34.70 & 76.06/6.38 & 54.16/31.81 & 72.88/9.96 & \textbf{88.03}/\textbf{3.07}\\ 
		\hline
		\textbf{Gibson}-Ribera & 44.29/33.27 & 79.26/6.53 & 81.21/5.74 & 88.62/4.13 & \textbf{95.69}/\textbf{1.15}\\ 
		\hline
		\textbf{Gibson}-Swormville & 58.81/18.10 & 55.46/22.19 & 45.43/33.78 & 66.86/13.43 & \textbf{92.58}/\textbf{1.68}\\ 
		\hline
		\textbf{Gibson-mean} & 45.80/34.48 & 68.30/14.42 & 63.30/21.09 & 80.45/7.44 & \textbf{92.10}/\textbf{2.83}\\ 
		\midrule
		\textbf{MP3D}-GdvgF & 68.45/11.67 & 81.78/5.48 & 82.39/5.14 & 80.99/5.69 & \textbf{91.05}/\textbf{3.93}\\ 
		\hline
		\textbf{MP3D}-gZ6f7 & 29.81/46.48 & 81.01/7.06 & 82.96/6.14  & 80.68/7.43 & \textbf{90.83}/\textbf{3.12}\\
		\hline
		\textbf{MP3D}-pLe4w & 32.92/30.79 & 66.09/12.78 & 66.76/11.82 & 76.41/8.03 & \textbf{96.20}/\textbf{1.68}\\ 
		\hline
		\textbf{MP3D}-YmJkq & 50.26/24.61 & 68.32/11.85 & 60.47/15.77 & 79.35/8.46 & \textbf{80.87}/\textbf{7.83}\\ 
		\hline
		\textbf{MP3D-mean} & 45.36/28.39 & 74.30/9.29 & 75.56/9.72 & 79.36/7.40 & \textbf{89.74}/\textbf{4.14}\\ 
		\bottomrule
	\end{tabular}}
\end{table*}
Denote the learnable parameters as $\vtheta=\{\alpha, \phi, \tau\}$, we can perturb the shallow decoder of $\gamma$ and $\phi$ to quantify the uncertainty regarding the scene geometry and appearance and adjust the planning strategy accordingly.
The training of networks takes a similar strategy with~\cite{Wang2023cvpr} to first penalize $l2$ errors of RGB and depth rendering:
\begin{equation}
  \mathcal{L}_{rgb}=\frac{1}{N} \sum_{n=1}^N (\hat{\vc}_n-\vc_n)^2, \mathcal{L}_{d}=\frac{1}{|R_d|} \sum_{r\in R_d} (\hat{d}_r-D[u,v])^2,
\label{eq:data_term}
\end{equation}
where the SDF values within and without the truncated areas are penalized as either the depth values or the truncated distance $tr$:
\begin{equation}
  \mathcal{L}_{sdf} = \frac {1}{|R_d|} \sum _{r\in R_d} \frac {1}{|S_r^{tr}|}\sum_{p \in S_r^{tr}} \big (s_p - (D[u,v] - d))^2,
\label{eq:sdf_term}
\end{equation}
\begin{equation}
  \mathcal{L}_{\text{free}} = \frac {1}{|R_d|} \sum _{r\in R_d} \frac {1}{|S_r^{\text{free}}|}\sum_{p \in S_r^{free}} (s_p - tr)^2,
\label{eq:free_term}
\end{equation}
and a smoothness term is performed to regularize the grid feature as:
\begin{equation}
    \mathcal{L}_{smooth} = \frac {1}{|\mathcal {G}|}\sum _{\vp \in \mathcal {G}} \Delta _x^2 + \Delta _y^2 + \Delta _z^2,
\end{equation}
where \(\Delta x, y, z = \mathcal{F}_\alpha(\vp + \epsilon_{x, y, z}) - \mathcal{F}_\alpha(\vp)\) refers to the featuremetric distances between adjacent samples on the hash-grid.

\section{Experiments}
\label{sec:experiments}
\subsection{Experimental setup}
The experiments are conducted on a desktop PC with an 
% Intel Core i9-12900K (16 cores @ 5.2 GHz), 64GB of RAM, and a single NVIDIA GeForce RTX 2080Ti.
Intel Core i9-10850K (10 cores @ 5.2 GHz), 64GB of RAM, and a single NVIDIA GeForce RTX 3090.
We utilizes the Habitat simulator~\cite{Habitat19iccv} with the Gibson~\cite{Gibson2018cvpr} and Matterport3D datasets~\cite{Matterport2017_3DV} for quantitative and qualitative evaluation. 
The agent observes posed RGB-D image sequence at a resolution of $256\times256$ and outputs discrete actions recursively, where the action space consists of \texttt{MOVE\_FORWARD} by $6.5$$cm$, \texttt{TURN\_LEFT} and \texttt{TURN\_RIGHT} by $10^{\circ}$, and \texttt{STOP}. The camera is set at a height of $1.25$m above the floor with a $90^{\circ}$ field of view vertically and horizontally.

%\noindent\textbf{Acc.}~($m^2$). The map accuracy metric evaluates the occupancy status that matches the ground truth in 2D space.
%
%\noindent\textbf{IoU}~(\%). The IoU metric evaluates the intersection ratio between the estimated map and the ground truth in 2D space. 

\begin{figure}
        \centering
	\subcaptionbox{Gibson-Elmira}[0.23\textwidth]
	{\includegraphics[width=0.25\textwidth]{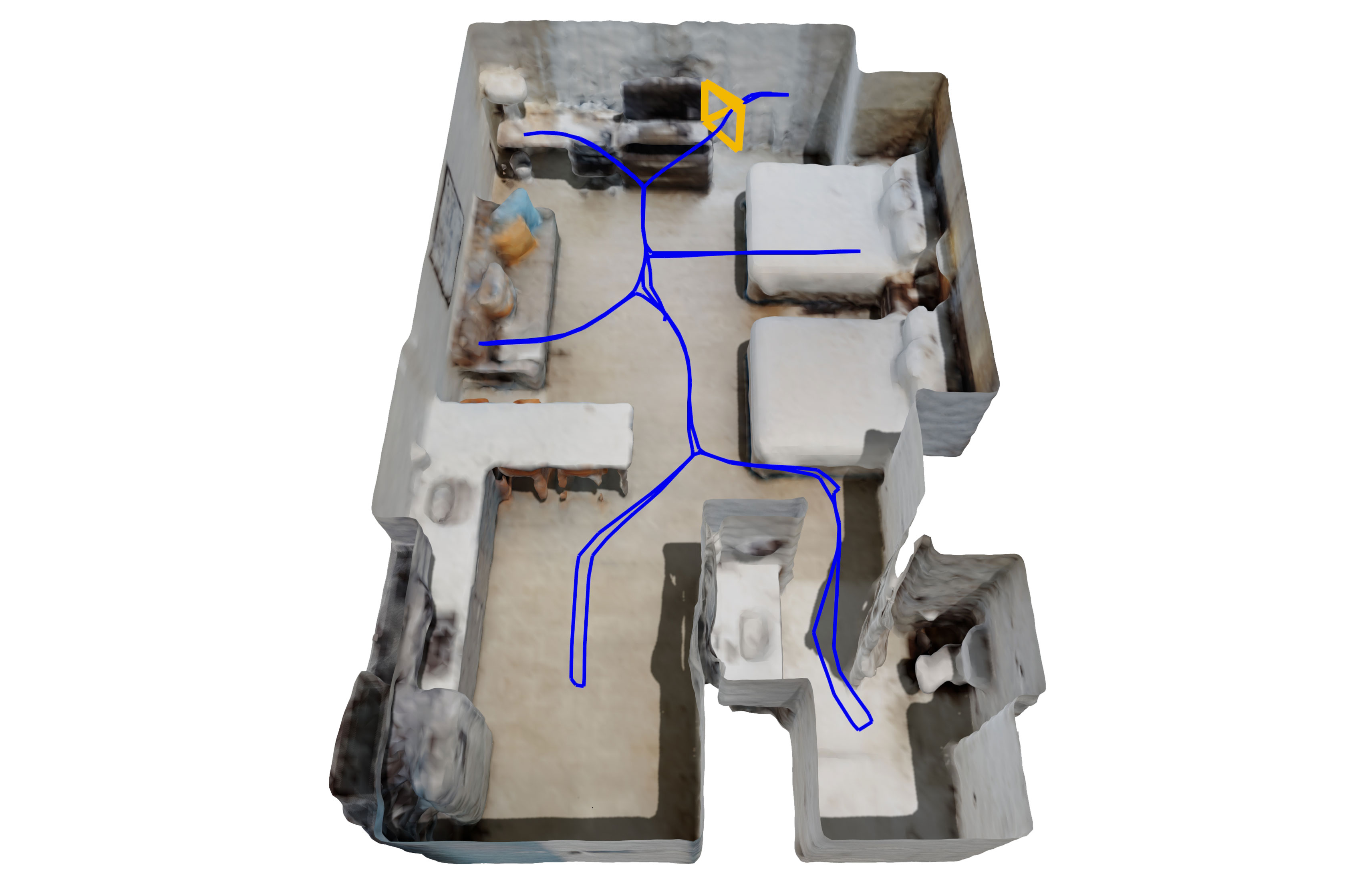}}
        \hfill
	\subcaptionbox{Gibson-Ribera}[0.23\textwidth]
	{\includegraphics[width=0.25\textwidth]{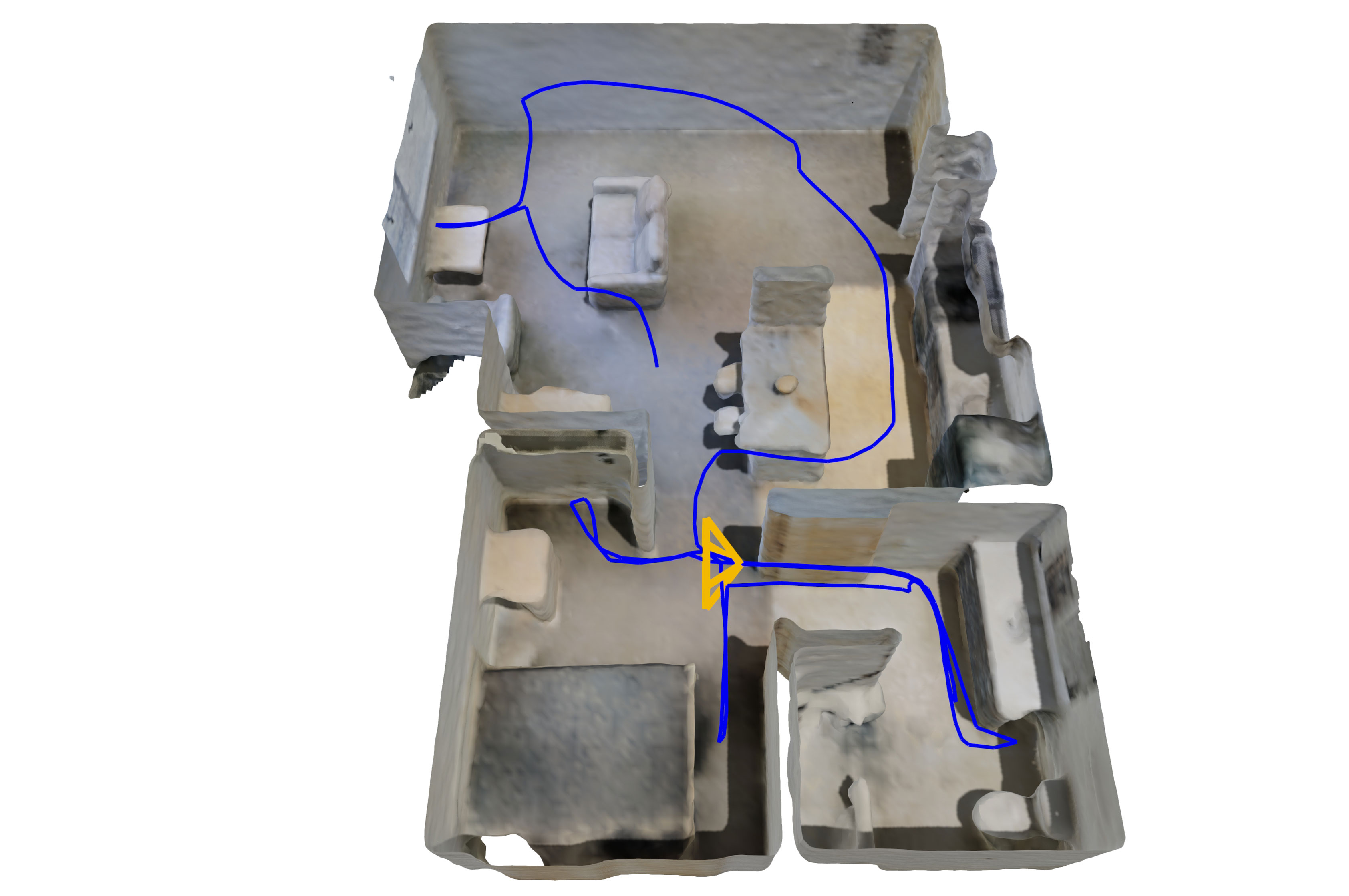}}\\
        \hfill
	\subcaptionbox{MP3D-gZ6f7}[0.23\textwidth]
	{\includegraphics[width=0.25\textwidth]{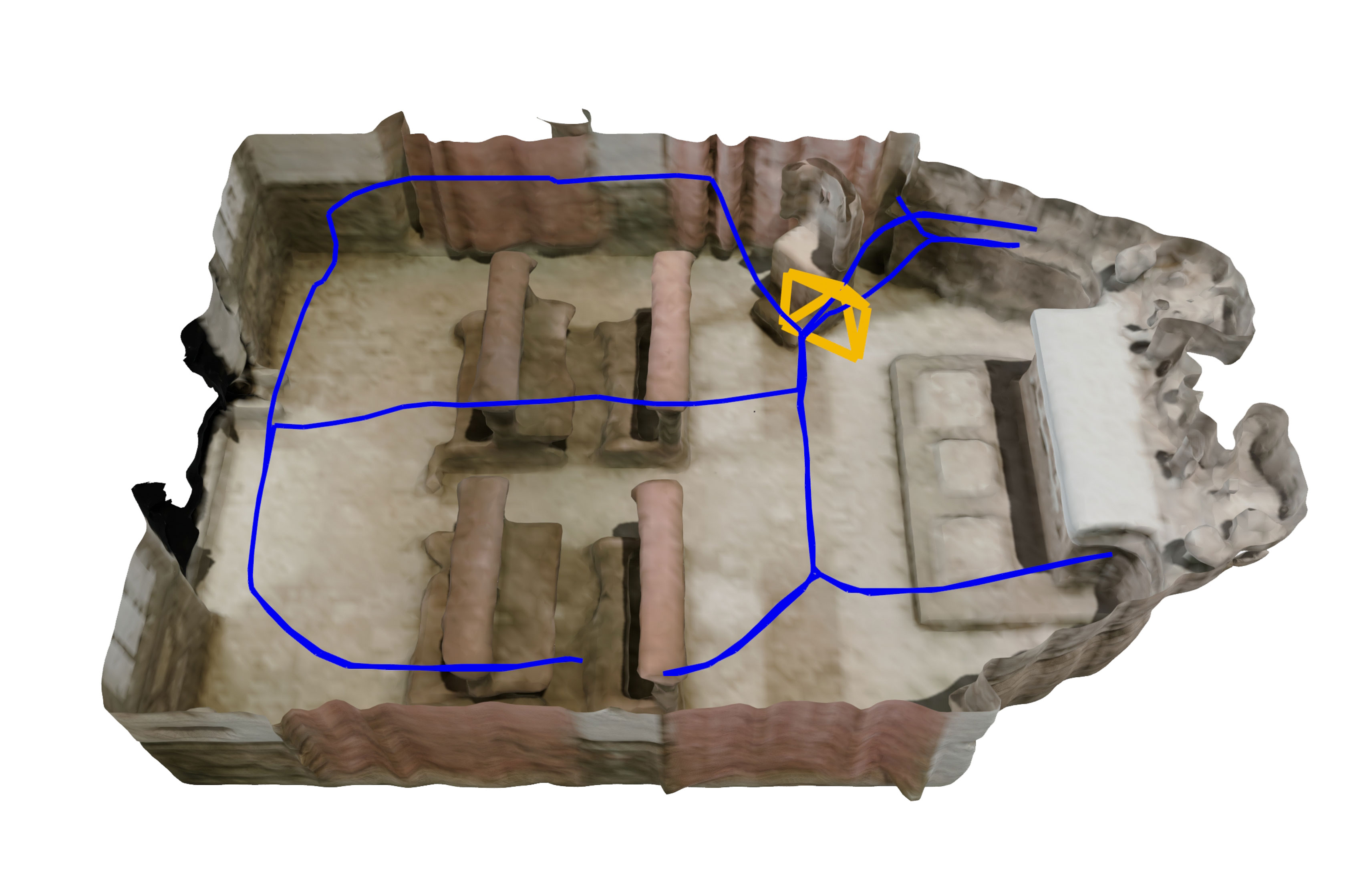}}
        \hfill
	\subcaptionbox{MP3D-pLe4w}[0.23\textwidth]
	{\includegraphics[width=0.25\textwidth]{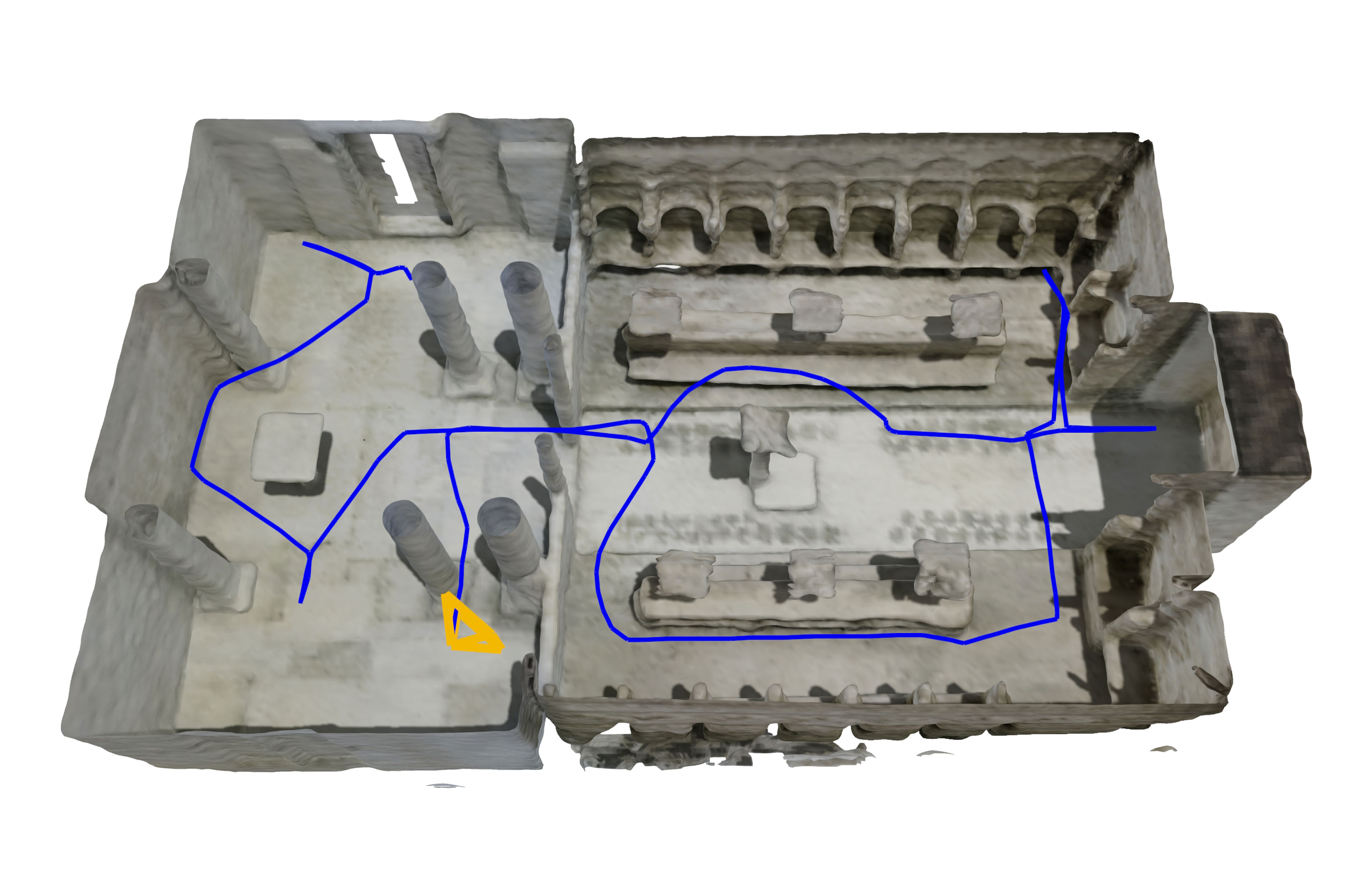}}
	\caption{The reconstruction results of small scenes through active exploration. }
	\label{fig:small_mesh} 
\end{figure}

\begin{figure}
	\centering
	\includegraphics[width=0.99\linewidth]{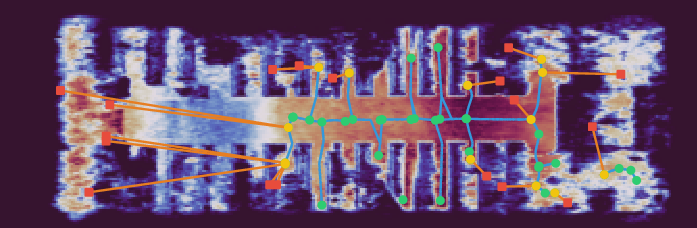}
	\caption{The complex scene with severe occlusions addresses challenges to trade-offs between efficient exploration and accurate reconstruction.}
	\label{fig:complex_scene}
\end{figure}

\begin{figure*}
        \centering
	\subcaptionbox{MP3D-Z6MFQ (22 rooms with 90.97\% of completeness)}[0.47\textwidth]
	{\includegraphics[width=0.5\textwidth]{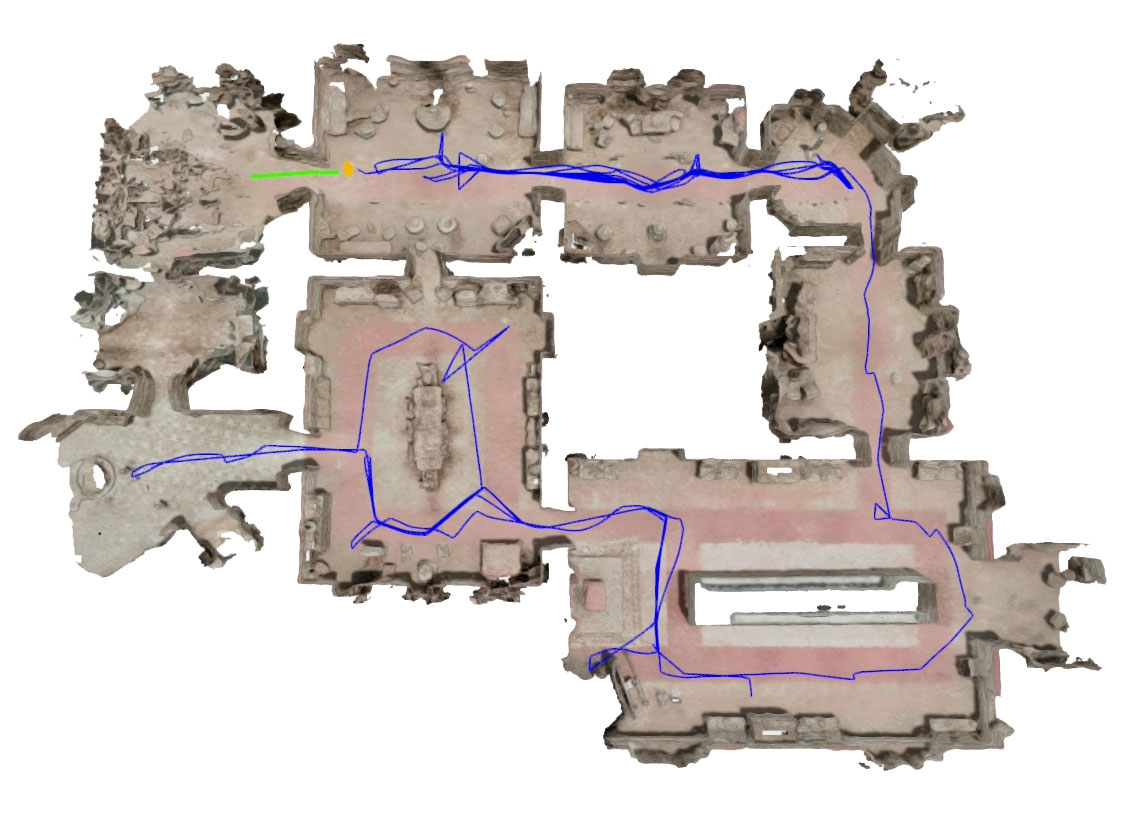}}
        \hfill
	\subcaptionbox{MP3D-q9vSo (22 rooms with 92.93\% of completeness)}[0.47\textwidth]
	{\includegraphics[width=0.5\textwidth]{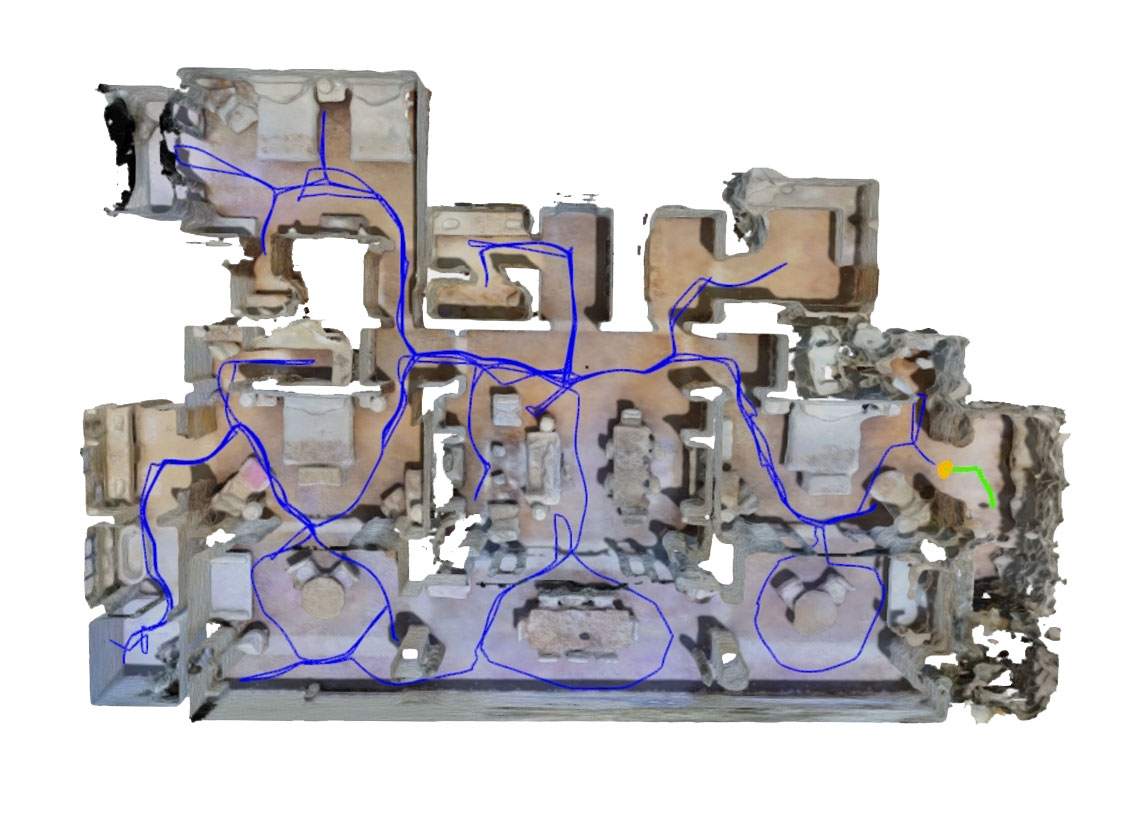}}
	\caption{The active reconstruction results of large-scale scenes of the Matterport3D dataset.}
	\label{fig:large_mesh} 
\end{figure*}

\begin{figure*}
        \centering
	\subcaptionbox{Random Voronoi vertex (64.78\%)}[0.32\textwidth]
	{\includegraphics[width=0.33\textwidth]{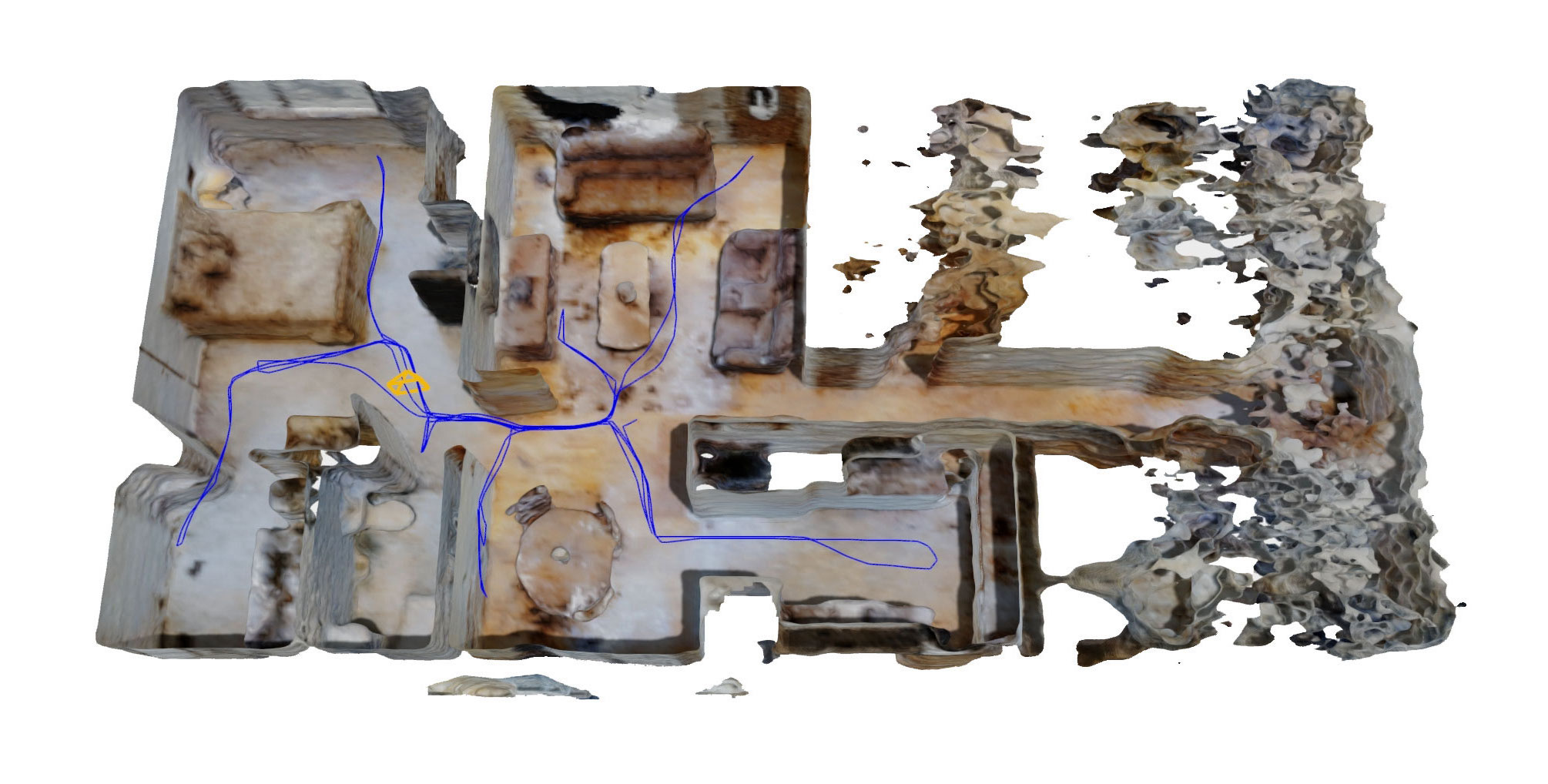}}
        \hfill
	\subcaptionbox{Best anchored vertex (73.93\%)}[0.32\textwidth]
	{\includegraphics[width=0.33\textwidth]{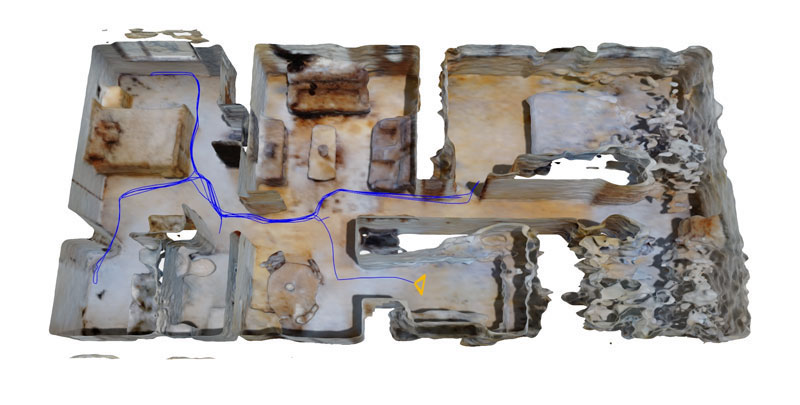}}
        \hfill
	\subcaptionbox{Ours (91.46\%)}[0.32\textwidth]
	{\includegraphics[width=0.33\textwidth]{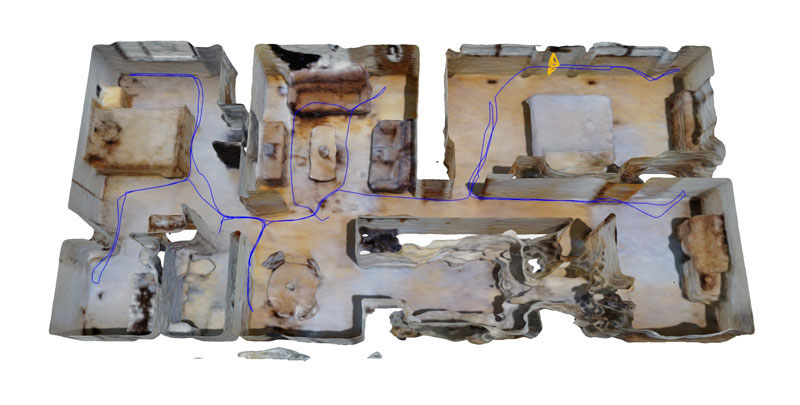}
	}
        \caption{The reconstruction results using different strategies of the next-best-view selection. (Gibson-Cantwell in 2000 steps)}
	\label{fig:ablation_target_vertex} 
\end{figure*}

\subsection{Comparisons against other methods}
We first follow the experimental setting of active neural mapping~\cite{Yan2023iccv} to evaluate the exploration performance on 13 diverse scenes. The quantitative evaluation is conducted using completion ratio (\%)\footnote{the percentage of points whose nearest distance is within 5cm} and completion ($cm$) metrics. 
As presented in~\cref{tab:method_compare}, the proposed active mapping system consistently outperforms the relevant methods of~FBE~\cite{Yamauchi1997frontier}, UPEN~\cite{Georgakis2022icra}, and ANM~\cite{Yan2023iccv}. The incorporation of the topology guarantees thorough exploration with promising reconstruction results (see~\ref{fig:small_mesh}). The system runs at 7-9HZ on average given different scales of scenes while achieving over 90\% of completeness with fine-grained details of geometry and appearance.
Nevertheless, as illustrated in~\cref{fig:complex_scene}, the complicated environment (MP3D-YmJkq) with severe occlusions and multiple narrow navigable pathways is still challenging. The exploration completeness is the worst among all tested scenes even though it still beats the competitors. 

\subsection{Reconstructing large-scale scenes}
Besides the test split in~\cite{Yan2023iccv}, we further validate the efficacy of the proposed method on the largest three scenes in test/val splits of Matterport3D dataset. Each scene consists of over 20 rooms.  As illustrated in~\cref{fig:large_mesh} and~\cref{fig:teaser}, the proposed method can achieve over 90\% completeness (including ceiling) and reconstruct fine-grained surface meshes in at most 10,000 steps. The integration of the topology map and the neural map assures thorough exploration and accurate reconstruction.
% \begin{table}[tb]
%   \caption{Reconstructed mesh quality on large-scale scenes.
%   }
%   \label{tab:large_scale}
%   \centering
%   \begin{tabular}{lccc}
%     \toprule
%      & \textbf{Rooms} & \textbf{Ours}\\
%     \midrule
%     \textbf{MP3D}-zsNo4  & 23 & 88.71/3.48\\
%     \textbf{MP3D}-Z6MFQ & 22 & 90.97/5.32\\
%     \textbf{MP3D}-q9vSo & 22 & 92.93/2.41\\
%     \textbf{MP3D-mean} &  & \\
%   \bottomrule
%   \end{tabular}
% \end{table}

\subsection{Ablation studies}
We conduct ablation studies to study the impact of different modules given the proposed strategies.
\paragraph{Voronoi vertex selection}
The proposed method takes account of both the accessible regions of interest and the visible regions of interest by anchoring the uncertain areas to the Voronoi vertices. We compare different strategies for the next-best-view selection. As illustrated in~\cref{fig:ablation_target_vertex}, randomly selecting the Voronoi vertex and filtering out previously visited ones lead to 64.78\% of completeness, while chasing the Voronoi vertex anchored by the most uncertain sub-areas leads to 73.93\% of completeness. The final result of ours prioritizes the anchored Voronoi vertex within the local horizon and follows a near-to-far order to traverse the vertices that have not been visited. The hierarchical framework leads to 91.46\% of completeness.

\begin{figure*}
        \centering
	\subcaptionbox{Without visibility guidance}[0.42\textwidth]
	{\includegraphics[width=0.46\textwidth]{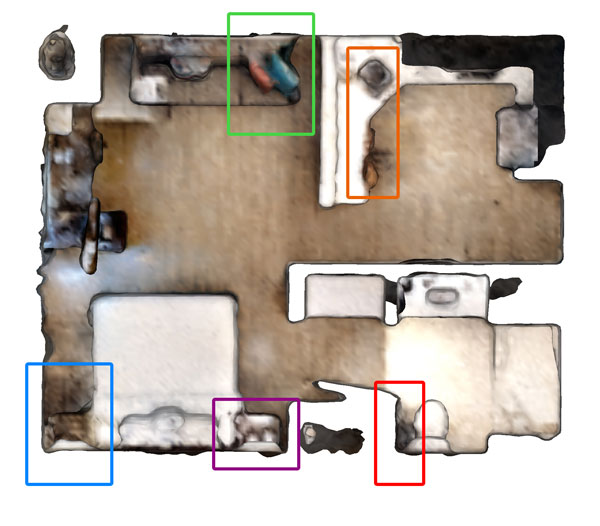}}
        \hfill
	\subcaptionbox{With visibility guidance}[0.42\textwidth]
	{\includegraphics[width=0.46\textwidth]{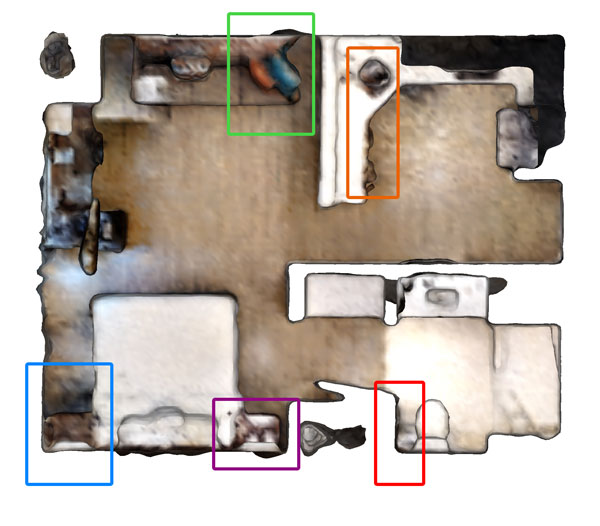}}
        \caption{Visibility guidance leads to the fine-grained reconstruction of details compared to accessibility-only exploration.}
	\label{fig:ablation_accessible_visible} 
\end{figure*}

\begin{figure}
\centering
    \includegraphics[width=0.99\linewidth]{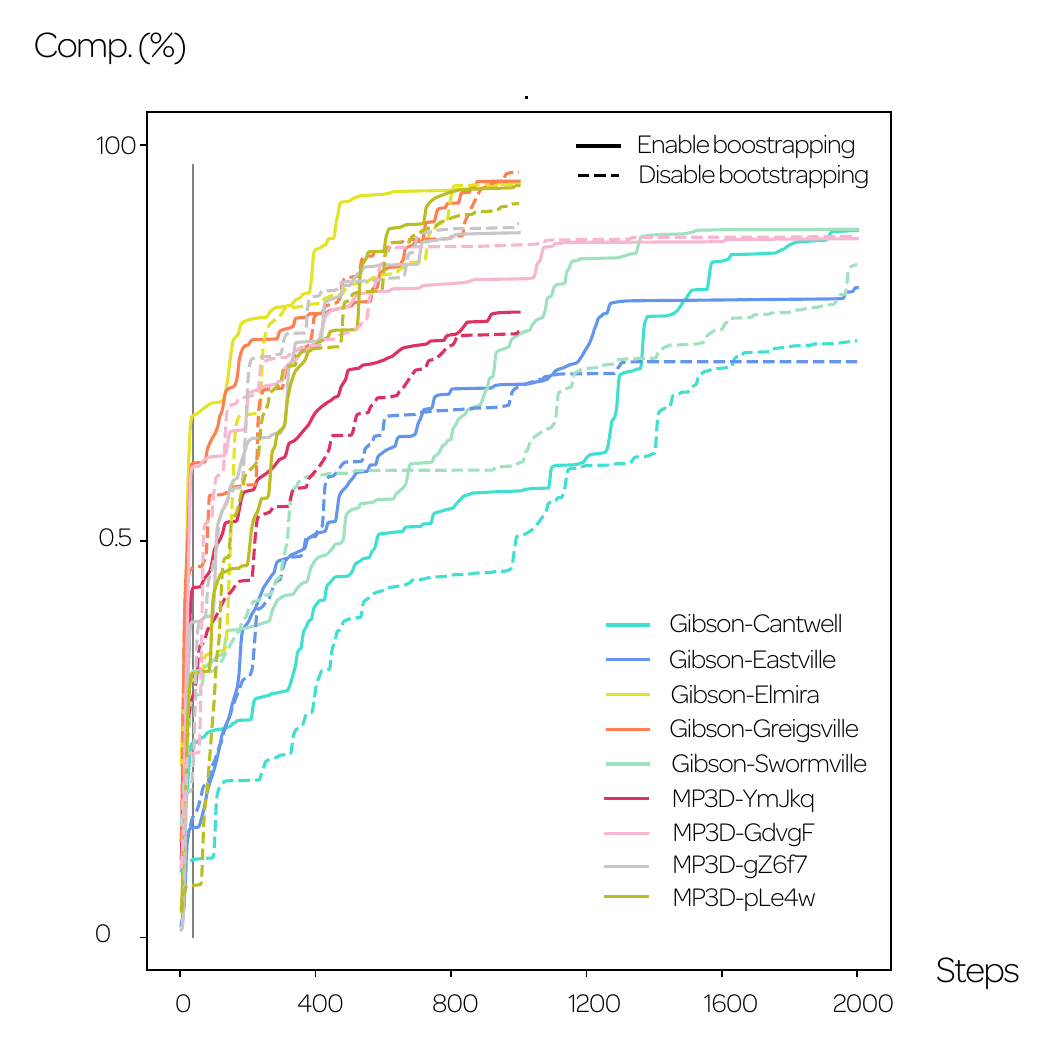}
	\caption{The bootstrap strategy (36 steps as the grey line for a complete rotation) leads to faster coverage at the beginning given a more confident initial state.}
	\label{fig:ablation_bootstrap}
\end{figure}
\paragraph{Visible regions of interest}
The proposed method targets an active neural mapping setting beyond pursuing the overall coverage of the environment. Therefore, we design delicate strategies concerning both local-global hierarchies and visible-accessible balance so the system can reconstruct accurate scenes with efficient traversal. As illustrated in~\cref{fig:ablation_accessible_visible}, we validate the efficacy of our visibility guidance from the uncertainty quantification method. The closeups validate that rotating towards the uncertain areas once approaching the Voronoi vertex leads to better details of geometry and appearance. The proposed method is designed to adopt a meticulous strategy near the target Voronoi vertex to search for unseen areas. This is due to the fact that the Voronoi vertex is the informative intersection of multiple pathways, where careful decisions should be made. The updated Voronoi graph on the other side defines the accessible regions of interest that guides the agent towards unvisited places while guaranteeing safe path planning during exploration. The collaboration of both sides achieves a nice trade-off between exploration efficiency and reconstruction accuracy.

\paragraph{Bootstrap strategy}
In the very beginning, we start by executing a complete rotation before the agent starts to explore. This strategy ensures a more confident understanding of the surrounding environment by sacrificing 36 steps. As is shown in~\cref{fig:ablation_bootstrap}, this simple strategy leads to faster ascendance of coverage in small scenes or when the agent is initialized at a complicated crossing. Starting with better confidence leads to consistently better completeness in most cases. The final completeness with or without bootstrap shows a similar ratio for all small scenes as the topology map enforces thorough exploration.

\section{CONCLUSION}
In this paper, we present a NeRF-based active mapping method. Taking advantage of the topology within the neural map and a hybrid network architecture, the proposed method greatly enhances the scalability of the active neural mapping problem. Taking visible and accessible regions of interest into consideration in a hierarchical framework, the proposed method achieves a nice trade-off between exploration efficiency and reconstruction accuracy with promising real-time capability. The experiments validate the state-of-the-art performance. Future potentials include the integration of more structural information about the environments such as semantics and relations between objects that better define the local similarity for efficient and accurate exploration and reconstruction.

\section{ACKNOWLEDGEMENTS}
The work is supported by NSFC (U22A2061, 62176010) and 230601GP0004.

% \addtolength{\textheight}{-12cm}  

%%%%%%%%%%%%%%%%%%%%%%%%%%%%%%%%%%%%%%%%%%%%%%%%%%%%%%%%%%%%%%%%%%%%%%%%%%%%%%%%

\bibliographystyle{IEEEtran}
\bibliography{main}

\end{document}